\documentclass{svproc}
\usepackage{amsmath,amssymb,amsfonts}
\usepackage{algorithm}
\usepackage{algpseudocode}
\usepackage{graphicx}
\usepackage{multirow}
\usepackage{subcaption}

\usepackage{url}

\begin{document}
\mainmatter        
\title{Transformer-based Neuro-Animator for Qualitative Simulation of Soft Body Movement}

%
%
\author{Somnuk Phon-Amnuaisuk}
\authorrunning{Phon-Amnuaisuk, S.}  
\tocauthor{}
\institute{Media Informatics Special Interest Group,\\
School of Computing and Informatics, Universiti Teknologi Brunei, \\
Gadong, BE1410, Brunei Darussalam \\
\email{somnuk.phonamnuaisuk@utb.edu.bn; span.amnuaisuk@gmail.com} }

\maketitle     

\begin{abstract}
The human mind effortlessly simulates the movements of objects governed by the laws of physics, such as a fluttering, or a waving flag under wind force, without understanding the underlying physics. This suggests that human cognition can predict the unfolding of physical events using an intuitive prediction process. This process might result from memory recall, yielding a qualitatively believable mental image, though it may not be exactly according to real-world physics. Drawing inspiration from the intriguing human ability to qualitatively visualize and describe dynamic events from past experiences without explicitly engaging in mathematical computations, this paper investigates the application of recent transformer architectures as a neuro-animator model. The visual transformer model is trained to predict flag motions at the \emph{t+1} time step, given information of previous motions from \emph{t-n} $\cdots$ \emph{t} time steps. The results show that the visual transformer-based architecture successfully learns temporal embedding of flag motions and produces reasonable quality simulations of flag waving under different wind forces.
\end{abstract}


\section{Introdcution}
Human cognitive abilities, particularly in conceptualizing and predicting the dynamics of physical events through vivid mental simulations, present a captivating area of exploration. One notable aspect of this cognitive capability is the capacity to engage in complex visual simulations without relying on explicit mathematical computations, as one would in a physics class. This phenomenon raises the fundamental question of how the human mind achieves such imaginative feats, prompting researchers to seek alternative approaches t qualitatively simulate physical activities.

In the context of dynamic tasks, such as describing the motion of soft-body objects like cloth influenced by external forces or the trajectory of rigid-body objects like a bouncing ball, humans effortlessly generate mental simulations. Importantly, this mental imagery does not involve precise mathematical calculations, creating an apparent gap between mathematical computation and visual imagination. This gap motivates the exploration of alternative approaches to qualitatively simulate physical activities within the AI community.

This study employs recent advancements in visual transformer models to replicate human-like visual imagination using artificial neural networks (ANNs). The visual transformer model undergoes supervised training using the movements of vertices on 2D flags exposed to wind forces. These movements in 3D space are generated through computationally intensive soft-body physical simulations, providing a comprehensive dataset for training the neural network. The transformer-based model is designed to capture the temporal context of the flag movement in 3D space influenced by wind and gravity, enabling it to predict the most probable consequences based on prior observations.

The decision to utilize a transformer-based architecture for this purpose marks a departure from the traditional application of transformers in natural language processing (NLP) domains. In contrast to the fixed vocabulary size assumed in NLP, by treating it as a fixed vocabulary-size classification problem, this study explores the adaptability of transformer architectures for soft-body simulations. The focus here is on dynamic visual sequence prediction, specifically predicting the positions of soft bodies in 3D space--a regression task.

Subsequent sections of this paper delve into the experimental methodology, providing a detailed account of the transformer model's architecture and elucidating the training process using simulated flag movements. The evaluation section assesses the model's proficiency in generating flag movements without relying on physical simulations, shedding light on potential parallels between neural network memory and the human capacity for visual imagination.

\section{Background}
The exploration of replicating human cognitive processes within the field of artificial intelligence (AI) has led to various research areas, with Qualitative Reasoning (QR) standing out prominently \cite{kuipers94}. QR in AI involves reasoning about the continuous aspects of the physical world using qualitative values, such as high, low, stable, increment, and decrement, without delving into precise numerical values. This qualitative approach is valuable in scenarios where parameters required for evaluating quantitative values are not available, or in scenarios where swift responses are needed, and decision-making relies on approximate, qualitative aspects \cite{trave03}.

\subsection{Qualitative Simulation}
One important aspect of QR is Qualitative Simulation (QSIM). QSIM operates on the premise of predicting outcomes from physical simulations without exact precision. Instead, QSIM provides an approximation of how outcomes might appear based on \emph{qualitative differential equations} (QDEs) of the system, which express states of continuous mechanisms based on incomplete knowledge of the system \cite{kuipers94}.

The mental model of flag motion can be modeled using QSIM by abstracting the mass, velocity, and other physical properties of the flag. If one wishes to visually present the movement, the abstraction in QSIM must also include relevant visual details (e.g., qualitative metrics of flag movement, velocity under a force, etc.). While exact precision in movement is not required, the computational process in QSIM can still be quite complex due to the need to capture these visual abstractions.

In this study, our qualitative simulation of flag movement adopts a machine learning (ML) approach. The ML approach relies on a qualitative understanding derived from representations shaped by experiences (i.e., training data). This approach offers an effective problem formulation. The knowledge representations in ML exhibit both static and dynamic properties: static information corresponds to memory snapshots, and dynamic information relates to transition models of these memory snapshots. Similar concepts have been explored for next-frame video prediction \cite{lee20,chang21}, providing an approximation that, while not pinpointing precise coordinates, still offers believable movement for human observers.

\subsection{Mental Models}
The concept that the brain constructs mental models was first articulated by Kenneth Craik in 1943 \cite{craik43}. While human mental models can simulate complex physical systems for prediction, the nature and operational characteristics of these models remain poorly understood. Early work in AI explored both symbolic approaches, such as \emph{STRIPS} \cite{fikes71}, and sub-symbolic approaches, such as \emph{TD-Gammon} \cite{tesauro95}, to construct models of observed environments.

In neuroscience and cognitive science, the term \emph{internal forward model} refers to a computational framework that predicts the outcome of actions or movements based on the current state of the system and past experiences \cite{kawato99}. Various computational frameworks have been explored to develop these models. For instance, early work by Grzeszczuk et al. \cite{grzeszczuk98} introduced the neuro-animator, constructed from a shallow artificial neural network model. Casey et al. \cite{casey21} proposed the animation transformer, which uses a transformer-based architecture to learn the spatial and visual relationships between segments across a sequence of images. Shannon et al. \cite{shannon21} developed a control model using \emph{process-description language} and \emph{genetic programming}. The explainability of these models remains elusive, prompting further investigation by many researchers. Phon-Amnuaisuk \cite{span17} investigated what was learned by a deep reinforcement learning model and suggested that trajectory of a ball (external physical mode) was captured in neural network weights when the agent was successfully trained to perform tasks. 

Battaglia and colleagues studied video game-based computer models and found that human physical reasoning aligns well with their computer model based on the Intuitive Physical Engine (IPE) framework \cite{battaglia13}. The IPE introduces probabilistic variation to standard deterministic physics simulations, showing that human judgments closely match predictions from these simulations rather than the actual physical world. To further promote studies in this area, Bakhtin et al. proposed a benchmark for physical reasoning named \emph{PHYRE} \cite{bakhtin19}.

\subsection{Adapting Transformer Architecture for Movement Prediction Tasks}
In the context of dynamic 3D-point arrays representing soft-body objects, the domain can be conceptualized as sequences of snapshots of the positions of points on a 2D flag. The desired mental model would accept sequences of snapshots and predict the next snapshot. Applying a visual transformer architecture to such dynamic sequences is a novel approach, adapting a technology initially designed for Natural Language Processing (NLP) tasks.

Adapting the transformer architecture, originally designed to handle text sequences for NLP, to the domain of dynamic 3D-point arrays presents several novel challenges and considerations. These challenges include: (i) handling sequences of 2D vertex frames, (ii) enabling next-frame predictions within the transformer framework, and (iii) obtaining embedding tokens for dynamic 3D-point arrays.

\subsubsection{Handling Sequences of Vertices Array}
Unlike text sequences, where spatial information is inherently captured in the order of text tokens, sequences of vertices arrays representing soft-body movement through time encode both temporal and spatial dimensions, adding an extra layer of complexity. In visual transformers \footnote{https://github.com/lucidrains/vit-pytorch}, 2D patches are assigned to each image frame, with each patch treated as a visual token, and spatial embeddings are applied to capture the contextual information within each patch. The same concept is extended to 3D vertices, where each patch along many frames forms a trajectory representing a 3D visual token. Extending 2D patches to 3D trajectories captures both temporal and spatial dependencies.

\subsubsection{Next-Frame Prediction with Transformer Architecture}
One of the primary objectives of our visual transformer model is to facilitate next-frame prediction, a task that inherently aligns with the dynamic nature of flag movement data. Achieving this within the transformer architecture requires careful consideration of how the model learns and represents temporal dependencies to predict the positions of vertices in a coherent and contextually relevant manner based on past movement.

\subsubsection{Embedding Tokens for Visual Vocabulary} 
In the realm of NLP, a fixed vocabulary size can be predetermined due to the discrete nature of textual data. However, the visual domain poses a different challenge: how to effectively obtain embedding tokens for a visual vocabulary. Unlike text, visual content is inherently continuous and lacks a predefined set of discrete symbols. In visual transformers, embeddings from visual input are not constrained by a fixed vocabulary.

\section{Problem Formulation}
\begin{figure}[hbt]
\centering
\includegraphics[scale=0.37]{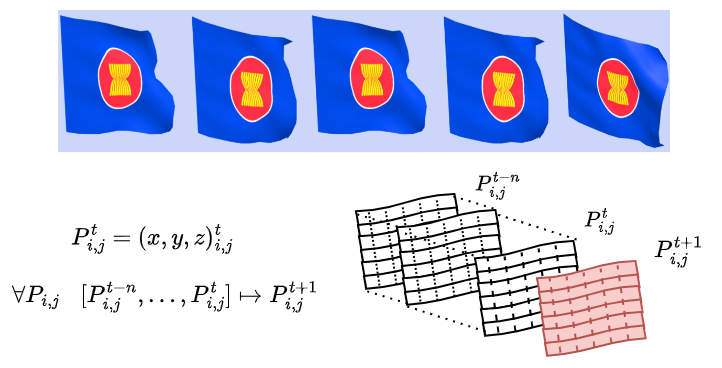} 
\caption{Segmenting a flag to 10 rows and 10 columns results in 121 intersection points. $P_{i,j}^t$ represents an intersection point (i.e., a particle) at row $i$ and column $j$ at time step $t$. A model can learn to map a sequence of snap shots to the next snap shot: $\forall P_{i,j} \;\;\; [P_{i,j}^{t-n},...,P_{i,j}^{t}] \mapsto P_{i,j}^{t+1}$ }
\label{flag1}
\end{figure}

Given a rectangular piece of soft body object (i.e., here is a flag) with the flag dimensions $i \times j$, where $i$ denotes the number of particles per row and $j$ the number of particles per column. The mechanical properties of the soft body object, such as \emph{stretching, shearing}, and \emph{bending}, can be effectively captured through a \emph{mass-spring model} (see \cite{va21}). Within this framework, each particle located at position $(x, y, z)$ and time $t$ is denoted as $P_{i, j}^t$. The temporal evolution of the flag is observed by tracking the positions of particles over time, forming a trajectory of length $n$ (see Fig. \ref{flag1}).

A visual transformer architecture \cite{alexey21} can be employed to model the spatio-temporal patterns of particles representing flag motions. It is noteworthy that the transformer architecture, originally designed for the natural language domain, relies on fixed-size tokens for input and output vocabulary. However, the continuous nature of particle movements in a 3D space in our setup introduces an additional layer of complexity when abstracting non-fixed size tokens representing these motions in a 3D space.

In our formulation, we address this challenge by considering each particle's trajectory as a token. The transformer model then maps an array of these trajectories to the next time step, denoted as $t+1$, based on observations from the previous $n$ time steps. Mathematically, this can be expressed as:
\begin{equation} \label{transformermapping}
\forall P_{i,j} \;\;\; [P_{i,j}^{t-n},...,P_{i,j}^{t}] \mapsto P_{i,j}^{t+1} 
\end{equation}

In this implementation, the flag is modeled as array of particles' positions oin 3D space with eleven rows and eleven columns, $P^{11 \times 11 \times 3}$. Considering 64 time steps then there are 121 trajectories, each with a length of 64. This forms an input with the shape of (64, 11, 11, 3). The functions \emph{trajectory\_embedder} and \emph{positional\_encoder} project the input into an embedding space with the shape (121, 128). Alg. \ref{transformer} outlines the transformer encoder model implemented here. The transformer based animator model outputs regression values representing positions at the next time step, $P_{i,j}^{t+1}$, according to equation (\ref{transformermapping}).

\begin{algorithm}
\caption{Transformer Based Neuro Animator}\label{transformer}
\begin{small}
\begin{algorithmic}[1]
\State \text{BATCH\_SIZE} = 32
\State \text{INPUT\_SHAPE} = (64, 11, 11, 3)
\State \text{NUM\_PATCHES} = 121
\State \text{LAYER\_NORM\_EPS} = 1e-6
\State \text{PROJECTION\_DIM} = 128
\State \text{NUM\_HEADS} = 8
\State \text{NUM\_TRANSFORMER\_LAYERS} = 8

\Function{transformer\_neuro\_animator}{trajectory\_embedder, positional\_encoder, input\_shape, num\_transformer\_layers, num\_heads, embed\_dim, layer\_norm\_eps}
    \State \textbf{Input:} \text{trajectory\_embedder}, \text{positional\_encoder}, \text{input\_shape},
    \State \quad \text{transformer\_layers}, \text{num\_heads}, \text{embed\_dim}, \text{layer\_norm\_eps}
    \State \textbf{Output:} \text{model}
    
    \State \text{inputs} $\gets$ \text{Input(shape=input\_shape)}
    \State \text{patches} $\gets$ \text{trajectory\_embedder(inputs)}
    \State \text{enc\_patches} $\gets$ \text{positional\_encoder(patches)}

    \For{iteration \text{{\bf in} range(num\_transformer\_layers):}}
        \State \text{x1} $\gets$ \text{LayerNormalization(epsilon=1e-6)(enc\_patches)}
        \State \text{attention\_output} $\gets$ \text{MultiHeadAttention(num\_heads=num\_heads,} 
        \State \quad \quad \text {key\_dim=embed\_dim // num\_heads, dropout=0.1)(x1, x1)}
        \State \text{x2} $\gets$ \text{Add()([attention\_output, enc\_patches])}
        \State \text{x3} $\gets$ \text{LayerNormalization(epsilon=1e-6)(x2)}
        \State \text{x3} $\gets$ \text{Sequential([}
        \State \quad \quad \text{Dense(units=embed\_dim * 4, activation=gelu),}
        \State \quad \quad \text{Dense(units=embed\_dim, activation=gelu)])(x3)}
        \State \text{enc\_patches} $\gets$ \text{layers.Add()([x3, x2])}
    \EndFor
    
    \State \text{representation} $\gets$ \text{LayerNormalization(epsilon=layer\_norm\_eps)(enc\_patches)}
    \State \text{representation} $\gets$ \text{GlobalAvgPool1D()(representation)}
    \State \text{representation} $\gets$ \text{Dense(units=11*11*3, activation="linear")(representation)}
    \State \text{outputs} $\gets$ \text{Reshape((1,11,11,3))(representation)}
    
    \State \text{model} $\gets$ \text{Model(inputs=inputs, outputs=outputs)}
    \State \textbf{return} \text{model}
\EndFunction
\end{algorithmic}
\end{small}
\end{algorithm}

\section{Transformer Regressor as a Neuro Animator}

\subsubsection*{Input/Output Representation for the Transformer Model}:
When observing the flag's configuration over $n$ discrete steps, the training data is constructed to predict the $(n+1)^{th}$ step, here n = 64. The training pair $(X,Y)$ is then formulated as $X^{64 \times 11 \times 11 \times 3} = P_{i,j}^{t-n}, \dots, P_{i,j}^{t}$ and $Y^{11 \times 11 \times 3} = P_{i,j}^{t+1}$. 

\begin{figure}[hbt]
\centering
\includegraphics[scale=0.53]{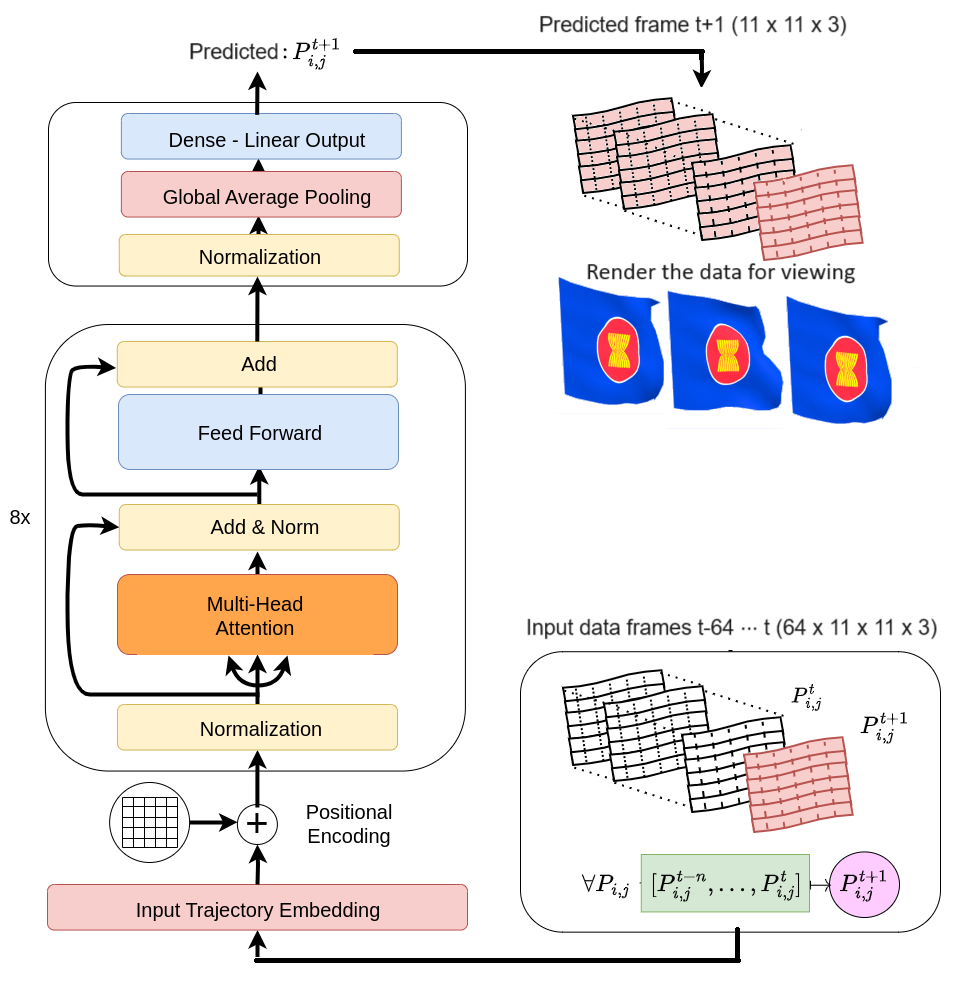} 
\caption{The transformer regressor model predicts the next positions of 121 points in the flag. Eight multi-head attention mechanisms were employed in each transformer layer, and the model implemented eight transformer layers. (see Alg. 1).}
\label{fig2}
\end{figure}

Here, $P_{i,j}^{t-n}, \dots, P_{i,j}^{t}$ represents the historical positions of the corner point $(i,j)$ up to $n$ steps, and $P_{i,j}^{t+1}$ denotes the target position for the subsequent time step.

The tensor $X$ was reshaped to a new tensor with size of (64, 121, 3). This forms 121 trajectories, each representing 64 time steps of a point in a flag in 3D space, allowing temporal information to be encoded in each trajectory. These trajectories then underwent the embedding process, resulting in an embedded representation of size (121, 128). Finally, each trajectory was encoded with positional encoding, which incorporates spatial information about the trajectories. The tensor $Y$ was reshaped to a new tensor with a size of (121, 3), forming the input/output for the transformer model (see Fig. \ref{fig2}).

\subsubsection*{Multi-Head Attention}: This implementation leverages the \emph{MultiHeadAttention} layer from \emph{Keras}. It takes three inputs, corresponding to \emph{Query, Key}, and \emph{Value}, (denoted as $Q, K$, and $V$, respectively). In this self attention case, all three inputs are set to the same input sequence. Eight multi-head attention mechanism were employed in this implementation, multiple attention heads enabled the model to attend to different parts of the input sequence simultaneously, allowing it to capture diverse patterns and relationships of the input sequence.

\begin{equation}
\begin{tabular}{l} 
$\text{head}_i = \text{Attention}(QW_{Qi}, KW_{Ki}, VW_{Vi})$ \\
$\text{MultiHead}(Q, K, V) = \text{Concat}(\text{head}_1, ...,\text{head}_h) W_O$
\end{tabular}
\end{equation}
where the input $Q$, $K$, and $V$ are projected into different spaces for each attention head using the weight matrices $W_{Qi}$, $W_{ki}$, and $W_{vi}$.
The learned weight matrix $W_O$ is used to project the concatenated outputs of all heads into the final output space.

\subsubsection*{Huber Loss}: In this implementation, the transformer regressor employed Huber loss. The Huber loss is a robust loss function that combines the best properties of mean squared error (MSE) and mean absolute error (MAE). 
\begin{equation}
\begin{tabular}{ll} 
$ L_{\delta}(y, f(x)) = \frac{1}{2}(y - f(x))^2 $ & $\text{if } |y - f(x)| \leq \delta$ \\
 $ \delta \left(|y - f(x)| - \frac{1}{2}\delta\right)$ & \text{otherwise}
\\
\end{tabular}
\end{equation}
where \(y\) is the true target value, \(f(x)\) is the predicted value by your model, and \(\delta\) is a hyperparameter that determines the point at which the loss transitions from quadratic to linear.

\section{Simulating Flag Motions by Prediction}
\subsection{Preparing Data Set}
Let $C$ be a piece of rectangular cloth (a soft body object representing a flag) with $w \times h$ particles, where there are $w$ particles per row and $h$ particles per column. The soft body object can be modeled using a \emph{mass-spring model} which captures the mechanical properties of stretching, shearing and bening of 
the frabic \footnote{https://en.wikipedia.org/wiki/Cloth\_modeling}. In a mass-spring model, each particle $p(i,j)$ where $i, j$ is the particle index, has a certain mass, and is connected to another particle with a \emph{spring} and a \emph{damper}.  It is also assumed that all particles have the same amount of mass and are subjected to two kind of system forces from spring ($F_s$) and damper ($F_d$):
\begin{equation}
F_s = K( L_0 - \|{\bf p-q}\|)\frac{({\bf p-q})}{\|{\bf p-q}\|} 
\end{equation}
where $K$ is the spring constant, ${\bf p}$ = $(p_x, p_y, p_z)$, ${\bf q}$ = $(q_x, q_y, q_z)$ are the positions of two particles; $L_0$ is the distance between two particles at rest;
\begin{equation}
F_d = -D{\bf v} 
\end{equation}
where $D$ is the damper constant, and ${\bf v}$ = $(v_x,v_y,v_z)$ is the velocity of the particle in 3D space. The system is also subjected to two kind of external forces from gravity ($F_g$) and wind ($F_w$):
\begin{equation}
F_g = (0, mg, 0), \mbox{   and   }
\end{equation}
\begin{equation}
F_w = (sin(w_x), 0, cos(w_z)) S_w
\end{equation}
where $g$ is the gravity, $m$ is the mass of each particle, $w_x, w_z$ are parameters to emulate an illusion of ripples on the flag cause by the wind, and $S_w$ is the strength factor that simulate breeze or strong wind.

In this implementation, we assume that a flag is fitted to a pole. Hence, particles at the edge on one side of the flag will not move due to any force. All other particles are influenced by the following forces in the system: $F_{s}, F_{d}, F_{g},$ and $F_{w}$. It is observed that this process gives a good visual approximation of cloth under the influence of different wind forces in real-time.

The dataset preparation process involves simulating the flag as described above and providing a rendering facility for visualization purposes. To limit the size of the data, we sample from 121 points from the flag. Each point, denoted as $P$, describes coordinates $(x, y, z)$, forming a 3D array with a shape of (11, 11, 3). 
The flag is simulated using three levels of wind strength, and the positions of points were recorded as training data. In this paper, we use the previous 64 states to predict the next state (see Fig. \ref{flag1} and equation (\ref{transformermapping})). This yields the data set with a tensor size of (15000, 64, 11, 11, 3).

\subsection{Results and Discussion}
Details of the transformer-based animator model, including the parameter settings and the overall process, are described in Algorithm 1 and Fig. \ref{fig2}. The model was built from scratch, leveraging Keras and the Transformer API. The \emph{Adam} optimizer and \emph{Huber} loss function were employed during training. The total number of trainable parameters was approximately 1.6 million after the model was compiled.

\begin{figure}[htbp]
\centering
\includegraphics[scale=0.26]{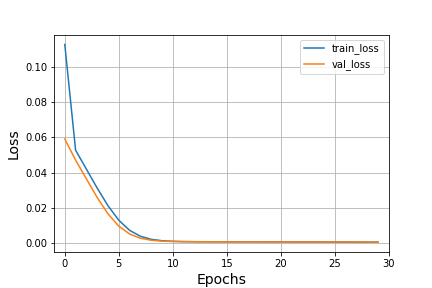}
\includegraphics[scale=0.26]{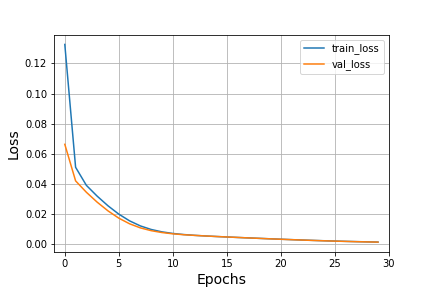}
\includegraphics[scale=0.26]{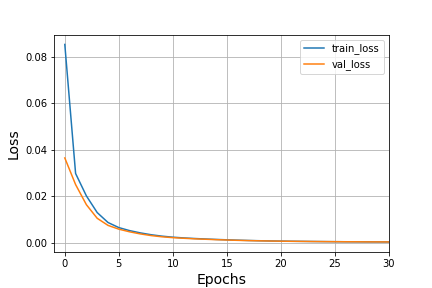}
\includegraphics[scale=0.36]{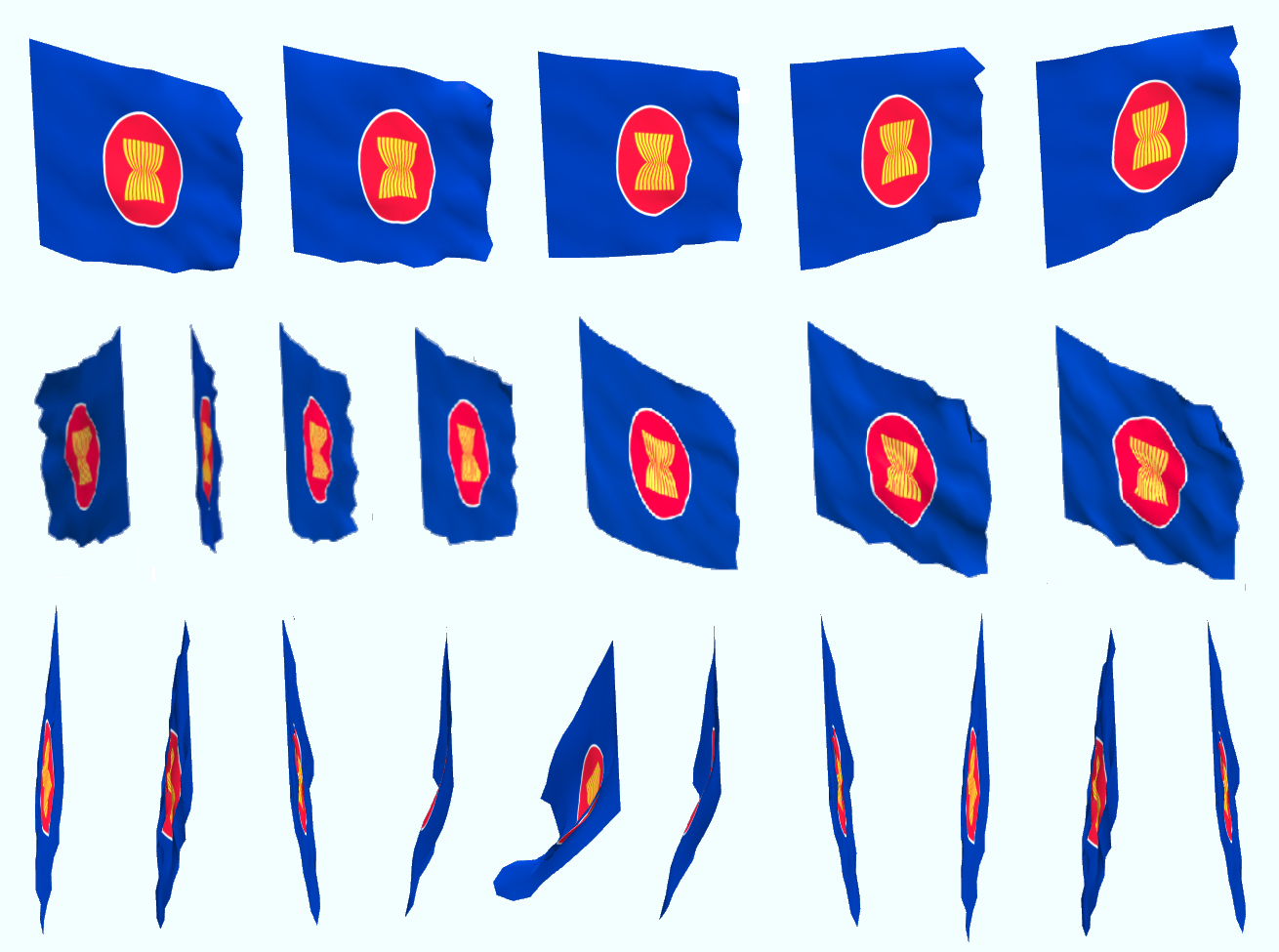}
\includegraphics[scale=0.38]{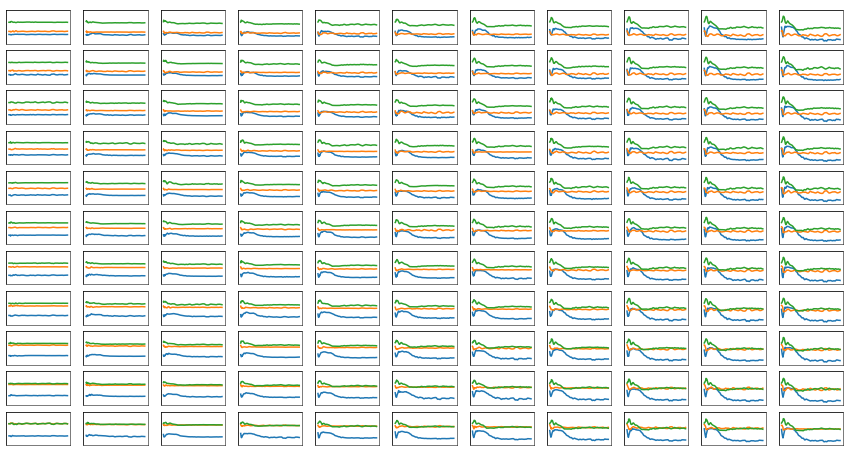}
\caption{Top pane: Training loss and validation loss from three neuro-animator models. Middle pane: Snapshots of predicted flag motion under three different wind strengths, from top to bottom: fluttering under strong wind, rippling under moderate wind, and hanging with occasional motions under light wind.
Bottom pane: Each of these 121 plots displays a particle $(x, y, z)$ positions over 1,000 frames. In each miniature plot, x-axis representing 1000 time steps and y-axis represent the particles' positions $(x, y, z)$ normalized to [-1, 1], $x$ is red, $y$ is green and $z$ is blue. }
\label{fig4}
\end{figure}

\subsubsection*{Qualitative Simulation as Prediction}
With the configured model, training data were fed into the model, which learned to make predictions based on the input-target pairs and adjusted its internal parameters to minimize the defined loss function. The predictions are points in 3D space that are rendered back as flag motions.

The trained model is used as an auto-regressive prediction model, where the current prediction is used as the input in the next time step. Three sets of flag motions under the influence of three different wind strengths are shown in Fig. \ref{fig4} 
(middle pane). The bottom pane of Fig. \ref{fig4} shows 121 plots. The plots in the leftmost columns are particles next to the fixed pole, hence particles' position $(x,y,z)$ are fixed. The plots in the rightmost columns are at the boundary and are much more active as expected.

To objectively quantify model performance,  frame prediction errors calculated from summation of discrepancies between targets and predicted points $| \hat P_{i,j}^{t+1} - P_{i,j}^{t+1} |$ are averaged over all test data (approximately 1000-1500 frames for each win strength) where $P_{i,j} \in [-1,1]$. The mean $\mu$ and standard deviation $\sigma$ are as follows: (i) strong wind: $\mu$ = 0.014, $\sigma$ = 0.005; (ii) moderate wind: $\mu$ = 0.025, $\sigma$ = 0.007; and (iii) no wind: $\mu$ = 0.018, $\sigma$ = 0.004.

\section{Conclusion}
This study draws inspiration from the innate human ability to mentally visualize dynamic events. It proposes that animation can be simulated based on memory recall from past experiences. Consequently, the model developed here does not focus on modeling control parameters for animating the 3D flag model but instead predicts the positions of the 3D flag directly.

We leverage the transformer model to learn the continuous nature of flag movement. The flag's spatio-temporal embeddings are captured using trajectory-based self-attention mechanisms within the transformer architecture. This approach enables the transformer model to learn and represent visually engaging movements, akin to the human mind's ability to simulate dynamic scenarios without explicit computational models, such as those commonly found in physics textbooks. Visual evaluations indicate that the generated flag motions appear believable. However, there is considerable room for improvement in terms of the naturalness of the motions (i.e., avoiding the uncanny valley). 


%
%


\begin{thebibliography}{6}
%
\bibitem{kuipers94}
Kuipers, B.: Qualitative Reasoning: Modeling and Simulation with Incomplete Knowledge. The MIT Press, Cambridge, Massachusetts (1994).

\bibitem{trave03}
Trave-Massuyes, L., Ironi, L., and Dague, P.: Mathematical foundations of qualitative reasoning. AI Magazine, 24(4), 91. (2003). https://doi.org/10.1609/aimag.v24i4.1733


\bibitem{lee20}
Lee, D., Oh, Y.J., and Lee, I.K.:
Future-frame prediction for fast-moving objects with motion blur.
Sensors, 20(16):4394 (2020). https://doi.org/10.3390/s20164394 

\bibitem{chang21}
Chang, Z., Zhang, X., Wang, S., Ma, S., and Ye, Y., Xiang, X., and Gao, W.:
MAU: A motion-aware unit for video prediction and beyond
Advances in Neural Information Processing Systems. 34, pp 26950--26962, Curran Associates, Inc., (2021).

\bibitem{craik43}
Craik, K. J. W.: 
The Nature of Explanation. Cambridge: Cambridge University Press (1943).

\bibitem{fikes71}
Fikes, R.E., and Nilsson, N.J.:
Strips: A new approach to the application of theorem proving to problem solving.
Artificial Intelligence, 2(3–4), 189-208, (1971).

\bibitem{tesauro95}
Tesauro, G.: Temporal difference learning and TD-Gammon. 
Communications. ACM, 38(3), 58-68, (1995).


\bibitem{kawato99}
Kawato M.:
Internal models for motor control and trajectory planning. 
Current Opinion in Neurobiology. Dec;9(6):718-27 (1999). doi: 10.1016/s0959-4388(99)00028-8. PMID: 10607637.

\bibitem{grzeszczuk98}
Grzeszczuk, R., Terzopoulos, D., and  Hinton, G.E.:
NeuroAnimator: Fast neural network emulation and control of physics-based models. 
In Proceedings of the 25th Annual Conference on Computer Graphics and Interactive Techniques, SIGGRAPH 1998, Orlando, FL, USA, July 19-24, 1998. ACM 1998, pp 9-20, ISBN 0-89791-999-8 (1998).

\bibitem{casey21}
Casey, E., P{\'{e}}rez, V., Li, Z., Teitelman, H., Boyajian, N., Pulver, T., Manh, M., and Grisaitis, W.:
The animation transformer: Visual correspondence via segment matching,
arXiv:2109.02614 [cs.CV] https://doi.org/10.48550/arXiv.2109.02614 (2021).

\bibitem{shannon21}
Shannon, P.D., Nehaniv, C.L., and Phon-Amnuaisuk, S.:
Cartpole Problem with PDL and GP Using Multi-objective Fitness Functions Differing in a Priori Knowledge. MIWAI 2021: 106-117 (2021).

\bibitem{span17}
Phon-Amnuaisuk, S.:
What does a policy network learn after mastering a pong game? MIWAI 2017: 213-222. LNAI Springer (2017).

\bibitem{battaglia13}
Battaglia, P.W., Hamrick, J.B., and Tenenbaum, J.B.: 
Simulation as an engine of physical scene understanding.
Proceedings of the National Academy of Sciences, 110(45):18327–18332, (2013).

\bibitem{bakhtin19}
Bakhtin, A., van der Maaten, L., Johnson, J., Gustafson, L., and Girshick, R.: Phyre: A new benchmark for physical reasoning. Advances in Neural Information Processing Systems, 32, (2019).


\bibitem{alexey21}
Dosovitskiy, A., Beyer, L., Kolesnikov, A., Weissenborn, D., Zhai, X., Unterthiner, T., and Dehghani, M., Minderer, M., Heigold, G., Gelly, S., Uszkoreit, J., and Houlsby, N.:
An image is worth 16x16 words: Transformers for image recognition at scale.
In the Proceedings of the International Conference on Learning Representations, (2021).

\bibitem{va21}
Va, H., Choi, M.H., Hong, M.:
Parallel cloth simulation using openGL shading language
Computer Systems Science and Engineering, 1. pp. 427-443, (2021).
doi:10.32604/csse.2022.020685
 

\end{thebibliography}
\end{document}